\documentclass{IEEEtran}

\usepackage{filecontents}
\usepackage{amsmath,amssymb,amsfonts}
\usepackage{algorithmic}
\usepackage{graphicx}
\usepackage{textcomp}
\usepackage{xcolor}

\begin{document}

\title{Structural Embeddings of Tools for Large Language Models\\}

\author{\IEEEauthorblockN{Eren Unlu} \\
\IEEEauthorblockA{\textit{Datategy SAS} \\
Paris, France \\
eren.unlu@datategy.fr}
}

\maketitle

\begin{abstract}

It is evident that the current state of Large Language Models (LLMs) necessitates the incorporation of external tools. The lack of straightforward algebraic and logical reasoning is well documented and prompted researchers to develop frameworks which allow LLMs to operate via external tools. The ontological nature of tool utilization for a specific task can be well formulated with a Directed Acyclic Graph (DAG). The central aim of the paper is to highlight the importance of graph based approaches to LLM-tool interaction in near future. We propose an exemplary framework to guide the orchestration of exponentially increasing numbers of external tools with LLMs, where objectives and functionalities of tools are graph encoded hierarchically. Assuming that textual segments of a Chain-of-Thought (CoT) can be imagined as a tool as defined here, the graph based framework can pave new avenues in that particular direction as well.

\end{abstract}


\section{Introduction}

Large Language Models (LLMs) have achieved remarkable success, progressively gaining a wider reputation of public \cite{brown2020language}\cite{wei2022emergent}. In spite of this positive recognition of emergent abilities, it did not take too much time for their weak points to be spotted. Many of these fallacies and incapabilities are well documented and studied by various researchers \cite{zhao2023survey}\cite{shuster2021retrieval}. Whilst exhibiting elevated performance in natural language understanding and composition, ironically, LLMs fail to possess very basic arithmetic reasoning, have simple temporal cognition or be aware of auto-fabricated pseudo-facts (hallucination), where much smaller specialized networks can excel \cite{bang2023multitask}\cite{imani2023mathprompter}\cite{dhingra2022time}.  

It can be discussed that estimating likelihood of the next token of a given text, which is the basis of LLM modeling, actually may be more “mimicking” the human conversational style rather than retaining most of its complex reasoning pathways, which also explains the perceived elevated performance of relatively smaller LLMs \cite{gudibande2023false}. In fact, we argue that the false promise of a sufficiently large generic transformer based architecture targeted to predict the next token to exhibit quasi-universal intelligence and a wide array of capabilities can be associated with the false promise of Universal Function Approximation of feed forward networks : Even though theoretically a sufficiently large deep neural network only composed of plain feed forward layers with non-linear activations should be able to perform any task on any type of flattened data. However, we know that in practice this is not possible, therefore we require specialized architectures with specialized formulations and loss definitions, such as convolutional networks for image or recurrent networks for sequential data. Therefore, it is not surprising for us to observe LLMs to struggle to execute mathematical operations on numerical tokens. Compartmented anatomy of the human brain gives also a glimpse on the importance of pre-designed specialized units for particular tasks \cite{kanwisher2010functional}. 

Given these observations, if current tensorial approach and homogenous transformer based architecture of LLMs will be preferred with increasing number of parameters, it is evident that incorporating external tools into LLMs’ context seem as the only viable solution to overcome aforementioned challenges. There have already been various attempts in the literature introducing interaction between LLMs and external agents \cite{komeili2021internet}. Actually, the idea of aligning conversational NLP models with external knowledge, primarily the internet, predates LLMs \cite{komeili2021internet}\cite{weston2014memory}. Works such as of \cite{gao2023pal} which enforce the LLMs to dismantle a given task into subtasks following the CoT principle, which are formulated as to be executed python scripts have demonstrated the importance of external agents. \cite{schick2023toolformer} is one of the most recent and prominent examples of a generic agent-LLM interaction framework. \cite{patil2023gorilla} presents an ambitious project to coordinate numerous different APIs with LLMs.

As explained above, if the current univariate neural architectural and implementational characteristics of LLMs persist in near future, developing much performant and generic frameworks to infuse external tools into language models is paramount of interest. One particular aspect of “tools” (any functionality which expects structured/quasi-structured inputs and produces structured/quasi-structured outputs) is their innate hierarchical and composable nature. In this paper, firstly we would like to highlight the importance of exploiting these characteristics of tools in the approaching era of augmented LLM-external agents synergy. Next, for this purpose, we provide the details of a hypothetical framework where the tools’ semantic definitions are encoded hierarchically using a specialized Graph Neural Network (GNN). The yielded embeddings respecting their DAG and recursive characteristics can be stored in vector databases and aligned with the queries intended to utilize them. Various other applications can be imagined based on this principle which may pave the way for further advancements. 

\section{Importance of Structural Embedding of External Tools}

Any external agent can be defined as a functionality which takes/produces a set of structured/quasi-structured input/output : $F(x)  \longrightarrow y$,  along with its semantic definition $D_{F}$, which is any textual description of the functionality, which may include also programming language syntax, structured scripts such as JSON dictionaries etc. For the sake of simplicity, we assume that this textual description is a single body of pure text which would fit into a single context of an NLP model and produces a single numerical vector as encodings. Note that, much more sophisticated approaches both for textual data and encoder modeling can be imagined for various purposes. Fig. 1. shows the illustration of a simple calculator which performs basic arithmetic operations on a list of given numbers, using python Read-Evaluate-Print-Loop (REPL).

\begin{figure}[h]
\label{fig:fig_1}
\includegraphics[width=0.99\linewidth]{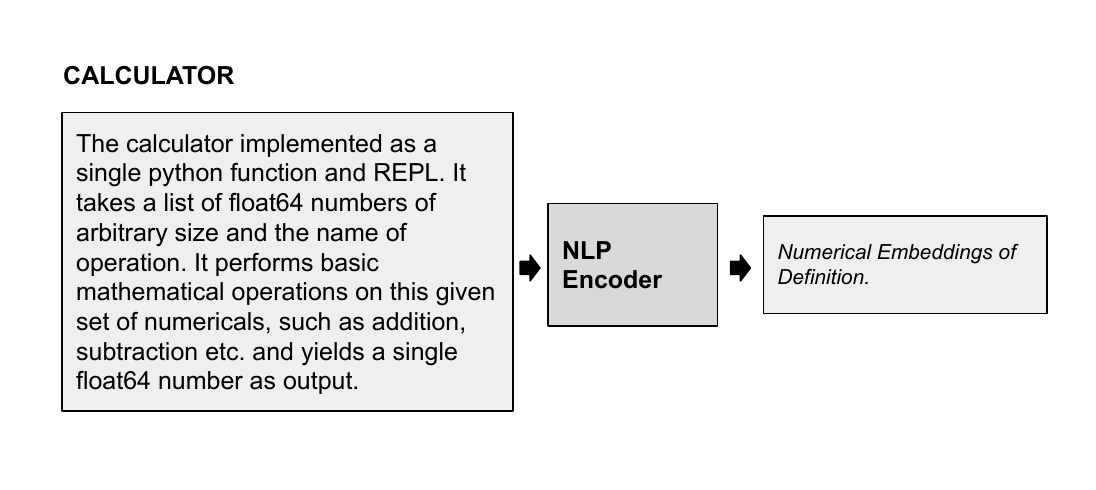}
\caption{The simplistic generic definition of a tool with a basic arithmetic calculator.}
\end{figure}

As mentioned previously, hierarchical and complex topological relations of functional ontology are ubiquitous. In the incoming era of synergy between LLMs and a massive number of external services, the capability to exploit this inherent nature to maximum will be of the utmost significance. First of all, to achieve this, the definition of functionalities (tools) shall be generic and universal as much as possible. Note that, any LLM based service can be considered as also an external agent, for another agent in this context. In order to highlight the importance of this phenomenon and compose the first guiding principles of a complex LLM centered multiagent ecosystem, in this paper, we propose a framework where textual descriptions of each agent is encoded into a fixed sized vector with a universal NLP model. It is important for the community using a framework such as that proposed, where each contribution and application is well documented, encoded, stored and shared. With this approach, we can exploit the complex and rich hierarchical, interconnected nature of a massive number of tools. Among important benefits we can list, which are not limited to these small number of examples : (1) For a given textual description of a task (For instance, a prompt in case of a pure LLM operation) one can perform augmented retrieval from vector databases by aligning the NLP embeddings of the task and the structural embeddings of agents. (2) Generative neural models can generate new tools from structural and query embedding pairs in the databases for given new tasks. Various other generative applications capitalizing on the hierarchical nature of tools can be imagined, probably incorporating graph neural modules up to a degree \cite{guo2022systematic}\cite{cheng2021molecular}. The generative framework can also fill the missing sub-components of a tool being developed. (3) Detailed complex analysis of ontology of tools.

In the next section, we present the principles of such a framework based on a specialized Graph Neural Network (GNN) and we underline various crucial aspects of to succeed.

\section{Hierarchical Graph Neural Network Embedding Framework of External Tools}

\subsection{Overview}

The central idea of this article is to present a pioneering framework where diverse participants of the community are developing external tools for LLM based systems. As mentioned previously, a tool or an agent can be defined and implemented in very divergent manners (different interfaces, different programming languages etc.), where they all refer to a basic, generic functionality, processing inputs coming from outside and ejecting outputs. Recursively, an LLM based operation can be defined as a tool itself. Based on the basic generic formulation , a tool is always represented by two set of textual definitions : First, the text describing the content and functionality of the tool, such as in Fig. 1. Second, prompt like queries conforming to human intentions, such as “What is 3 + 7 ?” in the case of the calculator example. The latter queries are optional for most of the components in the hierarchy, where they are required more to match tools with the intentions through vector databases.  

\begin{figure}[h]
\label{fig:fig_2}
\includegraphics[width=0.99\linewidth]{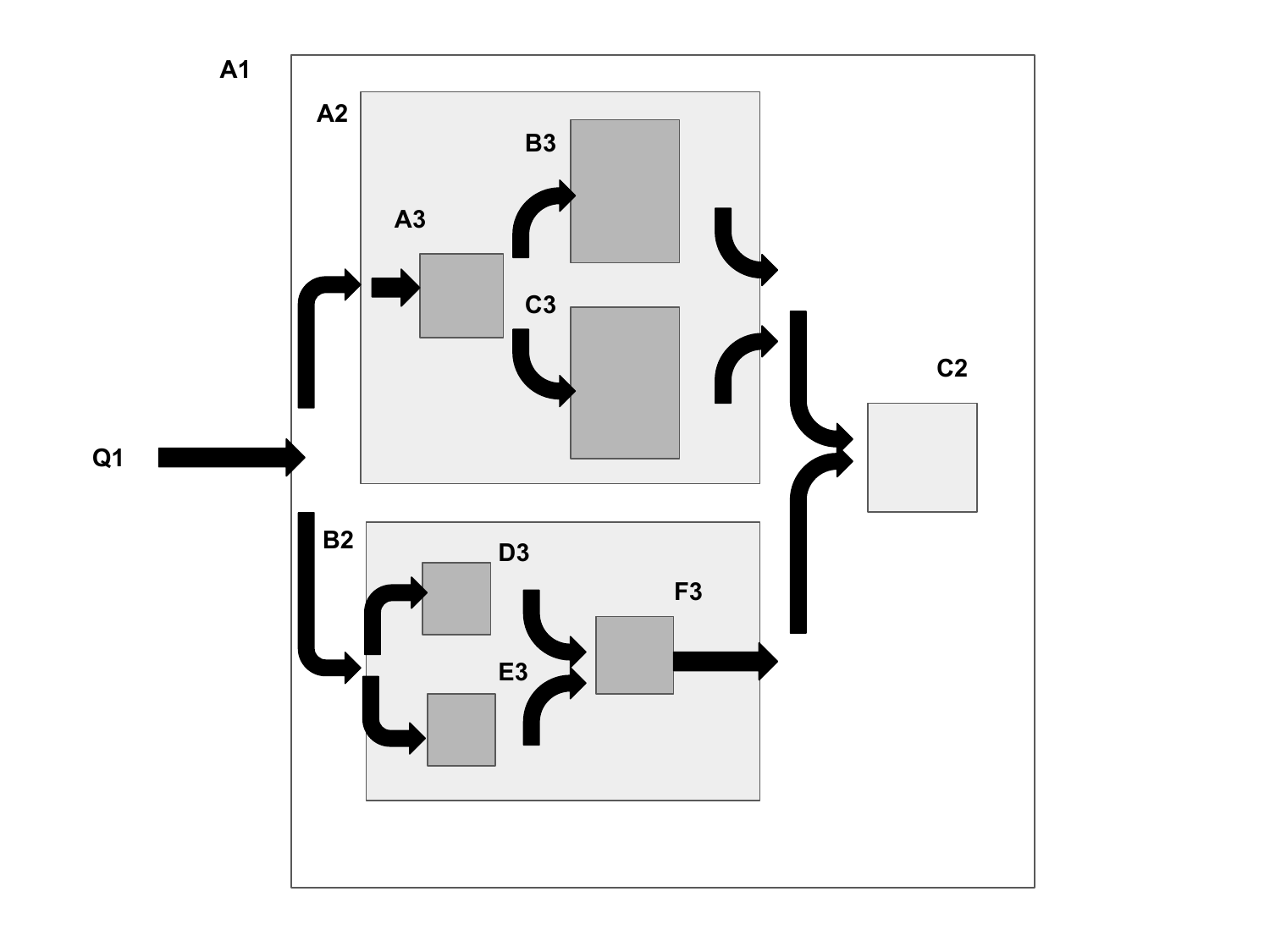}
\caption{Hierarchical graph structure of a simplified LLM based portfolio optimizer tool.}
\end{figure}

Fig. 2 illustrates a simplified hierarchical definition of a hypothetical LLM based tool (A1) which performs portfolio optimization on a given date with given parameters and criteria (Q1). This complex task of course requires a high number of interconnected hierarchical subtasks. In fact, in the real life implementation of such a framework, we expect tools to be designed much more in depth, which can go down to dozens of layers, most fundamental functionalities being at the bottom. For the sake of illustration Fig. 2 shows a simplified definition of a hypothetical agent. The names of the tools are written in an alphanumeric manner, where the number at the end defines the level of hierarchy starting from the top. As it can be seen, the portfolio optimizer is composed of 3 levels in hierarchy and 9 subtools in total. This imaginary automated portfolio optimizer fetches stock market data, public media news in textual form and macroeconomic indicators at the given date period in Q1. Its subtools are responsible of analyzing media for general market sentiment to compose a specific set of KPIs, check and alert if certain specific stocks are mentioned either positively or negatively in the news and finally constitute all intermediate indicators to an optimizer to produce the portfolio allocation. For instance, the simplified textual descriptions of several tools in the system can be listed as follows : 
\begin{itemize}
\item A1. “An LLM powered optimal portfolio optimizer. For the given date period, it fetches public media data from the predefined sources, hourly stock market data and daily macroeconomic indicators. An LLM with specific predetermined prompt templates produces a set of KPIs related to general market sentiment. Also, the public news are scanned to detect mentions of specific stocks. Later, these intermediate KPIs are gathered and an RL based portfolio optimizer produces the output.”
\item A2 : “An LLM powered public media analyzer. For the given period of dates and set of parameters, it fetches and normalizes the textual data using specific APIs. The voluminous data is processed in a multi-context fashion by referencing to embeddings in vector databases. First, it generates KPIs related to general market sentiment. Second, it generates a list of stocks mentioned in the media along with the KPIs related to their perceptions.”
\item B2 : “A python module which fetches hourly stock market data and macroeconomic indicators using a set of APIs for the given time period and produces a set of intrinsic KPIs.”
\item C2 : “A python module which takes the KPIs related to macroeconomics and stock market in a given time period and performs a portfolio optimization using reinforcement learning.”
\item A3 : “This component fetches and normalizes textual public data coming from newspapers, social media and similar sources, and implements complex NLP processes.”
\item B3 : “An LLM powered module which analyzes public media data, incorporates a complex transformer based NLP model which embeds voluminous data to embeddings in a vector database in chunks, then processes it with an LLM in a multicontext fashion, to generate KPIs related to overall market sentiment in given time period.”
\item C3 : “An LLM powered module which analyzes public media data, incorporates a complex transformer based NLP model which embeds voluminous data to embeddings in a vector database in chunks, then processes it with an LLM in a multicontext fashion, to generate list of KPIs related to stocks mentioned in the news as key-value pairs in given time period.”
\item D3 : “A component which takes a list of key-value pairs indicating public sentiment of multiple stocks and processes their stock market data purposefully, to generate new intermediate KPIs.”
\item E3 : “A component which fetches hourly open-close-low-high candlestick stock market data using APIs and generates intermediate KPIs for portfolio optimization algorithm.”
\item F3 : “A module that acquires intermediate KPIs related to general stock market data and highlighted specific stocks to generate further intrinsic features to feed the portfolio optimizer.”
\end{itemize}

This hypothetical example is probably a simplified version of a more complex real world example. In more practical applications we would expect more layers of hierarchy and number of submodules. However, still for certain types of other applications this level of abstraction may be preferred as well.

Using this given hypothetical example, we can detail our proposed framework of hierarchical graph embedding of tools. First of all, as we explained in the example itself, we expect a textual description of each submodule. For the sake of simplicity, we kept basic brief descriptions, but much more elaborated descriptions can be imagined. The codes or pseudo-algorithms of tools can be incorporated textually in the description, to further augment their perceptual capacity of embeddings. In addition to programming syntax or algorithmic pseudo-codes, structured representations can be involved both for expected input/output data form and the mechanism of action, such as JSON dictionaries. 

\subsection{Hierarchical Graph Neural Network Embedding of External Tools as Nodes}

The central idea of this article is to demonstrate a prototypical pioneering framework where diverse participants of a vibrant community participates by developing external tools for LLMs in a structured and hierarchical fashion as described in the previous section. Firstly, as mentioned previously, the textual descriptions are embedded into numerical representations by common identical NLP model(s) by contributors, which are publicly shared. The hierarchical DAG structure of subtools is precisely defined. This, in turn, makes every published tool a separate graph with its own set of nodes. With sufficient given number of tools in the central database, we can vest into Graph Neural Networks (GNNs) to capitalize on the richness of intertwined context. However, it is important that the chosen GNN respects the hierarchical and node anonymous nature of the problem. By node anonymity, we refer to the fact that it is much preferable not to encode identity indicators of nodes (such a name, ID field etc.) but totally encode the textual description. The advantages of such an approach are (1) similar components in the database would yield richer neural understandings at the end of the procedure. (2) an ID free, not overfitting representations can be produced and utilized (3) new tools can be generated resting on generative neural networks.

Ordinary GNNs such as vanilla Graph Convolutional Networks (GCNs) or Graph Attention Networks (GATs) do not conform to hierarchical structures, learning flat representations \cite{velickovic2017graph}\cite{zhang2019graph}. Inducing hierarchical knowledge in graph learning is well studied in the literature \cite{chen2021hierarchical}\cite{defferrard2016convolutional}\cite{cangea2018towards}\cite{ying2018hierarchical}. However, most of these propositions generate latent intermediate hierarchies out of unstructured data, possibly due to the nature of problems studied (generally node or graph classification or edge prediction). However, the problem we formulate in this study presents the hierarchical order of nodes (tools) well defined a priori. It is interesting to observe this type of a graph neural problem where hierarchical adjacency matrices are known in advance has not been explored sufficiently.   

\cite{sobolevsky2021hierarchical} proposes a neural network structure directly constructed based on the hierarchical structure or estimated latent hierarchy which in our case would be not suitable as each graph (tool) shall have different numbers of levels and nodes (sub-tools). We can state that the most analogous architecture to our formulation of the problem is \cite{fang2019hierarchical}. The authors propose to leverage on a hierarchical GNN to process multi-hop question answering, which can be found very similar to the CoT process in the context of LLMs. Their central problem also requires to pre-encode questions and answers with NLP models as node features and there exists an inherent hierarchical structure. 

Based on the constraints of the problem at hand, hierarchical graph encoding of external tools for LLM centered applications, we define the graph embedding framework as follows : (1) As each graph (tool) contains arbitrary number of layers and nodes, the framework shall adhere to this principle. (2) The loss function and optimization objective shall be properly defined where we do not have an obvious task such as edge prediction or graph classification a priori. For these purposes, in this paper we firstly propose a basic graph neural network architecture : A simple message passing algorithm where the main objective is defined as node representation learning, using a shared parameter encoder. 

Starting from the lowest layer in the main graph (tool) each subgraph is treated independently, where initial node features are embeddings produced by processing the textual description of the subtool. Using the traditional message passing in neighborhood and aggregation mechanism of certain GNNs, the objective is to predict the node feature of the parent node, which is also computed initially by encoding its textual description using the NLP model. This procedure is performed iteratively till reaching the top layer, the definition of main external tool. So, we can say it is formulated as a graph regression problem, where the regressions from the lower layers are propagated to parent nodes.

Assuming a database of many hierarchical external agent definitions contributed by the community, the proposed network can be trained efficiently to produce many resourceful applications. 

Fig. 3 depicts the message passing in the proposed framework for illustrative purposes, where 4 nodes on an intermediate layer l are arranged as in the figure, the output flow is connected to the parent external tool, as it is a proper node in the graph. As mentioned previously, including parent nodes, each node $j$ on layer $l$, $v_{l,j}$ has an initial node feature representation $f_{l,j}$, which is the embedding of its textual description by an advanced NLP model. In order to encode hierarchical structure further we include the encodings of layers as edge features, where the layer of the destination node is used. As it can be seen in Fig. 3, the node $v_{l,k}$ is connected directly to the parent node, therefore its edge feature is encoded based on layer $l+1$.

\begin{figure}[h]
\label{fig:fig_3}
\includegraphics[width=0.99\linewidth]{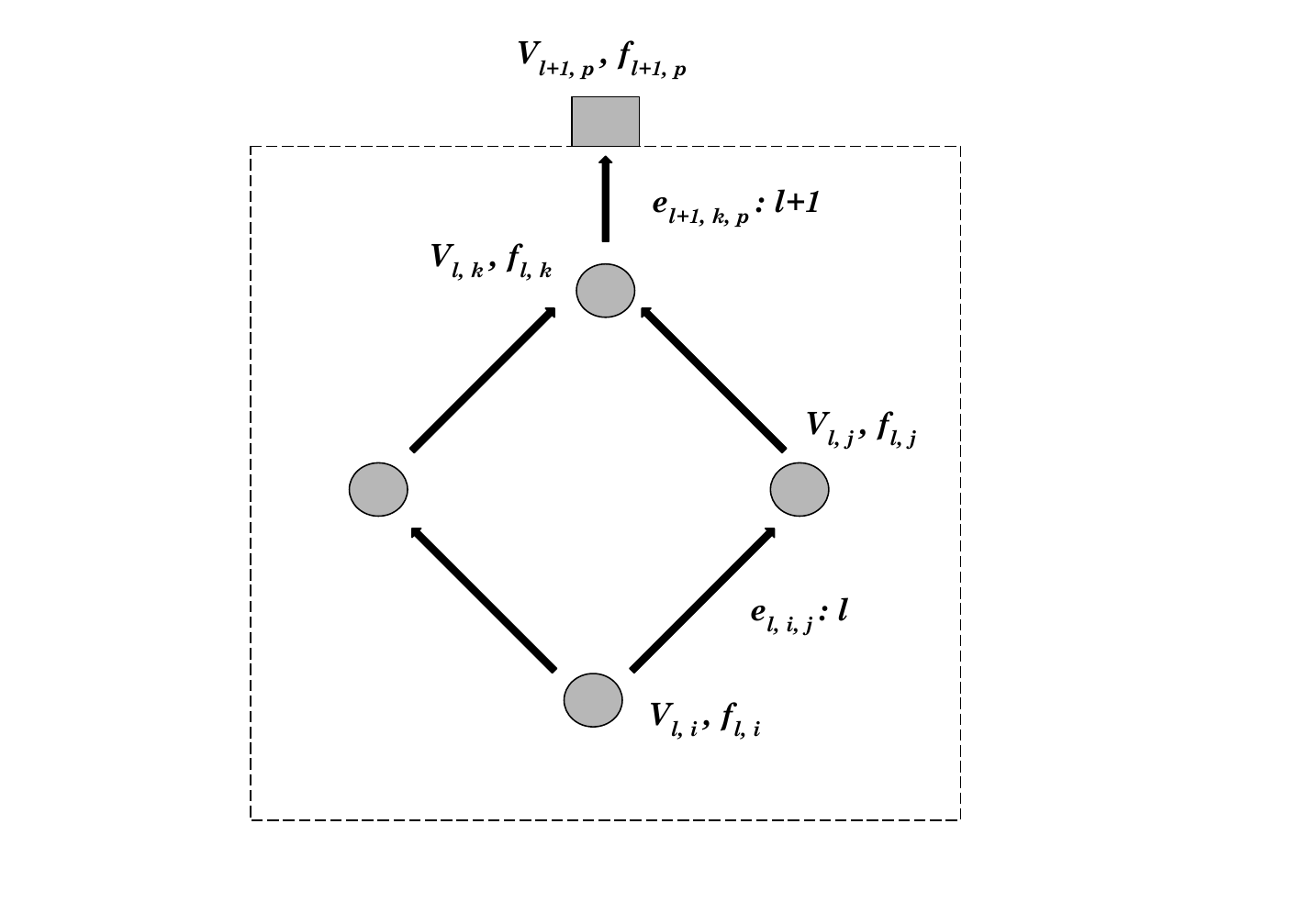}
\caption{Recursive hierarchical message passing in the proposed framework at a particular layer, simplifed exemplary illustration.}
\end{figure}

As mentioned previously, the deep learning objective is formulated as follows, starting from the lowest layer, $l=0$, using the standard GNN message passing the initial node representations of nodes are passed till reaching the parent node. The parent node also holds an initial representation of its textual description’s embedding and the loss function is defined based on regressing the parent node’s embedding. Following the nomenclature in \cite{ward2020practical}, we can write the equation as ($v', e'$ represent the features of nodes and edges respectively) :  

\begin{equation}
h_{l,i}^{(t)} = \sum_{j \in N_{l,i}}{f(v_{l,i}', e_{i,j}', v_{l,j}', h_{l,j}^{(t-1)})}
\end{equation}

Where $N_{l,i}$ denotes the set of vertices in the defined neighborhood of $v_{l,j}$, $v_{l,i}'$ denotes the node features of $v_{l,i}$, $e_{i,j}'$ denotes the edge features between $v_{l,i}$ and $v_{l,j}$, encoded as the proper hierarchical index as explained previously, $h_{l,i}^{(t)}$ denotes the latent embedding at iteration (level) $t$, $f$ is the shared neural network encoder with non-linearity, preferably a recurrent one due to intricacies of the studied problem (in case of a vanilla RNN for demonstration purposes) :

\begin{equation}
h^{(t)} = tanh(Wh^{(t-1)} + Ux + b)
\end{equation}

Using this message passing structure, at each hierarchy $l$, the $f$ is trained to minimize the regression error of the parent initial node feature $v_{l+1,p}$ :

\begin{equation}
argmin( \Vert v_{l+1,p}' - h_{l+1,p}  \Vert )
\end{equation}

The shared neural encoder is trained starting from the lowest layer upwards iteratively whilst all independent tools in the public database are trained together in batches. At the end of the procedure, where we have an able graph hierarchical neural encoder that one can use to produce rich node representations of tools and subtools. We can assume even if the objective is to regress on initial embeddings of the parent node, the latent output representations shall be much more plentiful than the initial NLP encodings which capture the hierarchical relationships.

The Chain-of-Thought (CoT) steps can be formulated and processed in this regard of the framework represented here, as the concept fits perfectly to it : Having an inherent hierarchical graph structure and node representations of textual encodings. This fact further enhances our belief in the study, that it shall be extended further by the community in the incoming age of abundant LLMs.

Note that, as we have stated initially in the paper, the further details of the neural architecture is out of scope of this study. For illustrative purposes a simple GNN architecture adhering to hierarchical principles is presented, however one can extend the framework with much more advanced architectures and formulations. 

\section{Conclusion and Perspectives}

Given the contemporary tensorial paradigm of LLMs, it is not hard to observe that a tendency will arise to incorporate a massive number of external agents orchestrated with multiple LLMs. To meet the demands of such an era, the hierarchical and ontological natures of tools should be well exploited. For this purpose, we propose a hierarchical GNN based framework where initial node features are encodings of textual descriptions of agents. Using a shared framework to encode numerous descriptions of tools in a public database, one can generate rich representations of tools capturing their complex intricacies. These rich encodings are then can be utilized for many applications such as automated tool generation, intelligent tool retrieval based on queries etc. This study presents a pioneering framework for such a framework and proposes a minimalistic architecture, where much more advanced formulations can be imagined.

\end{document}